%% file: main.tex
\documentclass[sigconf]{acmart}
\AtBeginDocument{%
  \providecommand\BibTeX{{%
    \normalfont B\kern-0.5em{\scshape i\kern-0.25em b}\kern-0.8em\TeX}}}

\setcopyright{acmcopyright}
\copyrightyear{2022}
\acmYear{2022}
\acmDOI{10.1145/1122445.1122456}

\acmConference[]{}{}{}
\acmBooktitle{}
\acmPrice{15.00}
\acmISBN{978-1-4503-XXXX-X/18/06}
\usepackage{multirow}
\usepackage{graphicx,booktabs,array}

\newcolumntype{C}{>{\centering\arraybackslash}m{6em}}

\usepackage{algorithm}
\usepackage{algpseudocode}

\newcounter{algsubstate}
\renewcommand{\thealgsubstate}{\alph{algsubstate}}
\newenvironment{algsubstates}
  {\setcounter{algsubstate}{0}%
   \renewcommand{\State}{%
     \stepcounter{algsubstate}%
     \Statex {\footnotesize\thealgsubstate:}\space}}
  {}
\begin{document}

\title{MultiBiSage: A Web-Scale Recommendation System Using Multiple Bipartite Graphs at Pinterest}

\author{
Saket Gurukar$^\dagger$, Nikil Pancha$^\ddagger$, Andrew Zhai$^\ddagger$, Eric Kim$^\ddagger$, Samson Hu$^\ddagger$, \\ Srinivasan Parthasarathy$^\dagger$, Charles Rosenberg$^\ddagger$, Jure Leskovec$^\ddagger$ }

\affiliation{%
	\institution{$\dagger$ The Ohio State University, $\ddagger$ Pinterest }
	\country{United States}
}
\email{gurukar.1@osu.edu, {npancha, andrew, erickim, sansonhu}@pinterest.com}
\email{ srini@cse.ohio-state.edu, {crosenberg, jure}@pinterest.com, }

\newcommand{\xhdr}[1]{{\noindent\bfseries #1}.}
\renewcommand{\shortauthors}{Gurukar et al.}
\newcommand{\jure}[1]{{{\textcolor{red}{[Jure: #1]}}}}

\begin{abstract}
Graph Convolutional Networks (GCN) can efficiently integrate graph structure and node features to learn high-quality node embeddings. These embeddings can then be used for several tasks such as recommendation and search. At Pinterest, we have developed and deployed PinSage, a data-efficient GCN that learns pin embeddings from the Pin-Board graph. The Pin-Board graph contains pin and board entities and the graph captures the pin belongs to a board interaction. However, there exist several entities at Pinterest such as users, idea pins, creators, and there exist heterogeneous interactions among these entities such as add-to-cart, follow, long-click. 

In this work, we show that training deep learning models on graphs that captures these diverse interactions would result in learning higher-quality pin embeddings than training PinSage on only the Pin-Board graph. To that end, we model the diverse entities and their diverse interactions through multiple bipartite graphs and propose a novel data-efficient MultiBiSage model. MultiBiSage can capture the graph structure of multiple bipartite graphs to learn high-quality pin embeddings. We take this pragmatic approach as it allows us to utilize the existing infrastructure developed at Pinterest -- such as Pixie system \cite{eksombatchai2018pixie} that can perform optimized random-walks on billion node graphs, along with existing training and deployment workflows. We train MultiBiSage on six bipartite graphs including our Pin-Board graph. Our offline metrics show that MultiBiSage significantly outperforms the deployed latest version of PinSage on multiple user engagement metrics.

\end{abstract}


\maketitle

\section{Introduction}
\input{introduction}

\section{Preliminaries and Related Work}
\input{related_work}


\section{Methodology}

\input{methodology}

\section{Experiments}
\input{experiments}

\section{Conclusion}
\input{conclusion}

\bibliographystyle{ACM-Reference-Format}
\bibliography{references}

\newpage
\input{appendix}
\end{document}

%% file: introduction.tex
Pinterest helps its users discover things they love by using 100+ billion pins present on its platform. The users discover things by exploring the personalized recommendations that are powered by Graph Convolutional Network (GCN) based recommendation system, PinSage\cite{ying2018graph}. PinSage aggregates the visual and textual features of pins along with their local graph neighborhood from the Pin-Board graph to generate the pin embeddings. The learned pin embeddings are then fed as input to several machine learning models at Pinterest for personalized recommendation, spam classification, ads recommendation, and search.

\begin{figure}[t]
    \centering
    \includegraphics[width=0.95\linewidth]{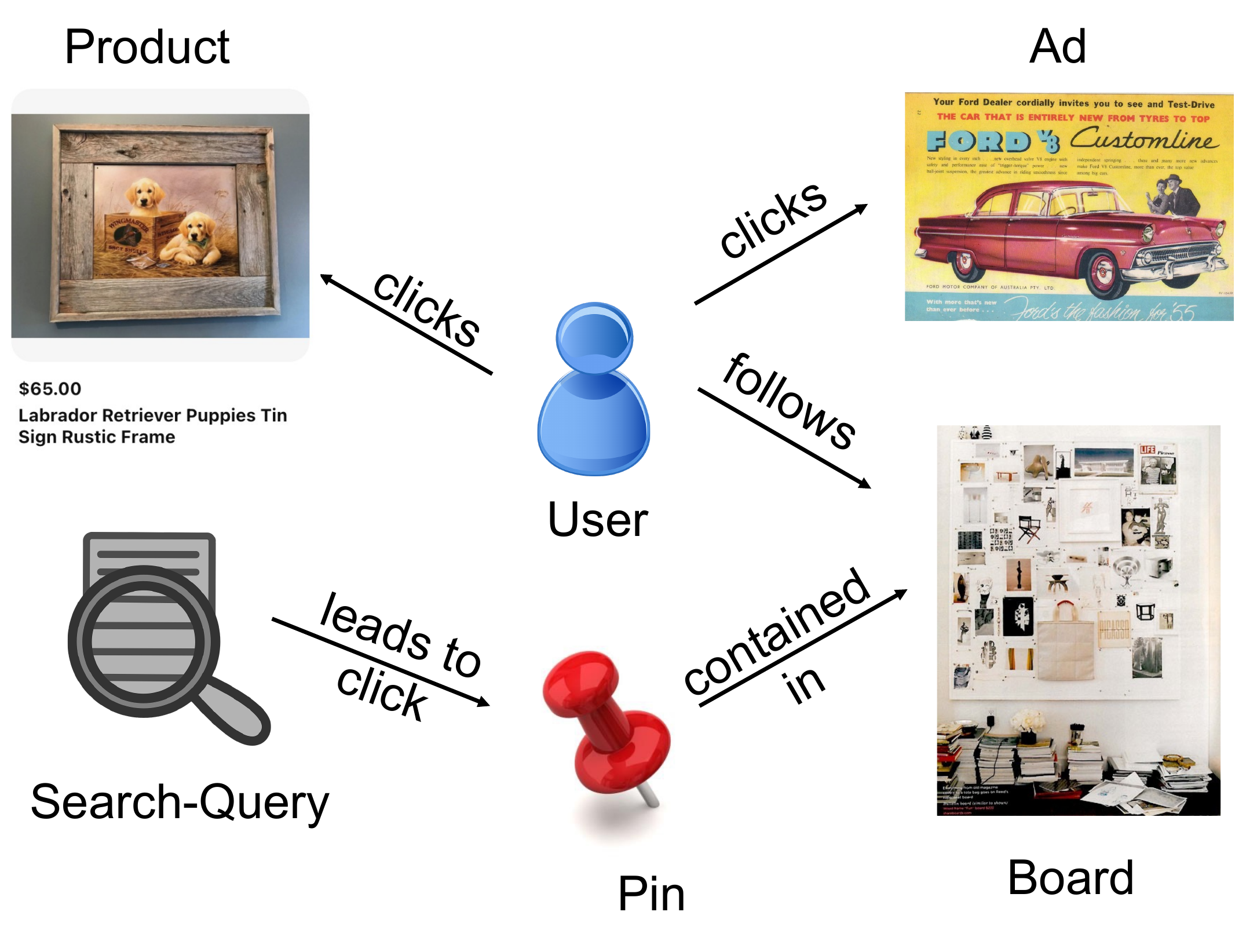}
    \caption{Pinterest's Heterogeneous Graph}
    \label{fig:pin_graph}
\end{figure}
The PinSage \cite{ying2018graph} model currently relies on only the Pin-Board bipartite graph to identify the local graph neighborhood of the pin. However, at Pinterest, we have multiple entities such as users, idea-pins, creators, products, ads, search-queries, boards, along with other entities. Moreover, these entities can have diverse interactions on Pinterest. Figure \ref{fig:pin_graph} shows an example heterogeneous graph at Pinterest. Here, a user can click on a product or an ad, a user can follow a board, a pin can belong to a board, a user can enter a search-query and then click on a pin shown in the search results. We hypothesize that these interactions among entities contain rich signals that can be utilized to learn high-quality pin embeddings.

Recently, several extensions of Graph Convolutional Networks for heterogeneous graphs have been proposed \cite{chang2015heterogeneous, zhang2019shne, tang2015pte, jiang2020task, hu2020heterogeneous}. These proposed models often operate on moderate-size to large-scale graphs. One such heterogeneous graph model, HGT \cite{hu2020heterogeneous}, showed promising results on Open Academic Graph  \cite{zhang2019oag} consisting of 179 million nodes and 2 billion edges. However, the proposed version of HGT is trained on a single machine and requires all the training data (heterogeneous graph and node features) to be present on that single system. At Pinterest, in an arbitrary version of our graph, we have 2+ billion pins, 2+ billion boards, 7+ billion pin-board edges and 400+ million users \cite{userstat}. 
The storage of graph and node features itself requires more than 3TB of space. Hence, given the scale of the data at Pinterest, training existing heterogeneous graph models on a single machine is not feasible. There also exist a few frameworks such as Distributed Deep Graph Library (DistDGL) \cite{zheng2020distdgl} and Aligraph \cite{zhu2019aligraph} that support distributed training of heterogeneous GNNs models. DistDGL and Aligraph trained distributed GCNs models on Ogbn-papers100m \cite{hu2020ogb} ($\sim$111 million nodes) and Taobao-large ($\sim$483 million nodes) graphs, respectively. Both DistDGL and Aligraph rely on multi-processing for "on-the-fly" sampling of node's neighbors. On a 4+ billion node graph, the "on-the-fly" sampling of neighbors in every epoch or after a few epochs, results in a high training time of the model. The running time of sampling can be decreased by increasing the number of workers but that results in additional costs. 
Hence, it is challenging to train the heterogeneous graph models on web-scale graphs.

\textbf{Present work}:  In this work, we adopt a pragmatic approach that allows us to utilize the existing infrastructure developed at Pinterest to train heterogeneous graph models. The existing infrastructure includes the random-walker system, Pixie \cite{eksombatchai2018pixie}, that loads the graph (without node features) in a single AWS x1.32xlarge (2 TB RAM, 128 cpus) instance and utilizes C++ routines to identify the local graph neighborhood of the nodes. We also utilize the existing model training and feature extraction workflows. The approach can be summarized as follows. Given a heterogeneous graph we first decompose it into multiple bipartite graphs. We then compute the local graph neighborhood of pins with the help of Pixie. To efficiently aggregate the node features and node neighbor features from multiple bipartite graphs, we propose a transformer based MultiBiSage model. Given $k$ bipartite graphs, MultiBiSage learns $k$ pin embeddings with the help of transformer \cite{vaswani2017attention} model where each pin embedding corresponds to each input bipartite graphs. These pin embeddings are then aggregated with another transformer layer that generates the final pin embedding. 
The contributions of our work are summarized as follows
\begin{enumerate}
    \item To the best of our knowledge, this is the largest-ever application of  heterogeneous graph models on  web-scale graphs with over 4.8+ billion nodes and 9.7+ billion edges. 
    \item We show that our proposed MultiBiSage model significantly outperforms the currently deployed PinSage model.
    \item We present several ablations studies and case study that helped us arrive at the design of MultiBiSage. 
\end{enumerate}

%% file: related_work.tex
In this section, we present the necessary definitions and provide a review of related work that utilize graph convolutional networks for recommendations. We also review distributed training frameworks for heterogeneous graph models.

\subsection{Definitions}

\begin{definition}{Heterogeneous Graph:} A heterogeneous graph $H=(V, E, T, R, X, \phi, \psi, \varphi)$ is a graph where $V,E, T, R$ are the set of nodes, set of edges, set of node types, and set of relations, respectively. A node $v\in V$ has node type $\phi(v): V \rightarrow T$ and an edge $e (u,v) \in E$ between two nodes $u,v$ has edge type $\psi(e(u,v)): E \rightarrow R$. A node $v$ can have node features denoted by $ \varphi(v): V \rightarrow X$.
\end{definition}

Heterogeneous graphs can naturally capture multi-modal interactions in the real-world. For instance, nodes can represent diverse entities such as users, pins, boards, products, and ads; while edges can represent diverse  interactions types such as click, add-to-cart, and purchase.

\begin{definition}{Bipartite Graph:} A bipartite graph $B=(V_1, V_2, E, X, \varphi)$ is a graph consisting of two sets of nodes $V_1$ and $V_2$ and an edge $e(u,v) \in E$ between two nodes $u\in V_1, v \in V_2$ has one edge type. A node $v$ can have node features denoted by $ \varphi(v): V \rightarrow X$.
\end{definition}

\begin{definition}{Heterogeneous Graph Embedding}: Given a heterogeneous graph $H$, a heterogeneous graph embedding method learns a function $f(H): V \rightarrow \mathbb{R}^d$ such that $\Theta = f(H)$ is a matrix consisting of $d$-dimensional node embeddings. The function $f$ is learned such that similarity between nodes in the heterogeneous graph is approximated by the closeness between nodes in the embedding space.
\end{definition}

In this work, we aim to learn a embedding function $f'$ such that 
\begin{equation}
    f(H) \approx f' (\cup_{V_i} \cup_{V_j}  \cup_{E_k} \; B(V_i, V_j, E_k, X, \varphi))
\end{equation}
as computing $f(H)$ might not always be feasible.

\subsection{Related Work}
\subsubsection{Proximity-based heterogeneous graph models} Given a heterogeneous graph, the proximity preserving heterogeneous graph models rely on the notion of similarity between nodes in the graph space to learn the node embedding. The similarity in the graph space can be defined through random-walks visits \cite{grover2016node2vec} or first/second-order proximity between nodes \cite{tang2015line}. Metapath2vec \cite{dong2017metapath2vec} is a popular random-walk-based model that learns the embeddings of nodes in the heterogeneous graph by performing random walks through meta-path schema. Metapath2vec inspired several meta-path schema-based heterogeneous graph models. For instance, HIN2Vec \cite{fu2017hin2vec} learns node embeddings by estimating the presence of meta-path between two nodes. Meta-path schema weighting techniques are also proposed \cite{chen2017task, sun2013pathselclus} to learn efficient node embeddings. Other random-walk-based approaches, proposed models that do not require specifying meta-paths. For instance, JUST \cite{hussein2018meta} proposed a Jump and Stay random-walk strategy while Hao et al.\cite{hao2021walking} proposed an attention mechanism for performing random-walk on heterogeneous graphs. The first/second-order proximity-based methods define proximity based on the edge-weights or the number of times the nodes are part of a subgraph or meta-graphs \cite{yang2020heterogeneous}. PTE \cite{tang2015pte} decomposes a heterogeneous graph into multiple bipartite graphs and learns node embeddings by preserving the first-order and second-order proximity between nodes in each bipartite graph. The PTE model is trained through the proposed joint training framework. PTE is considered as a shallow network embedding method as it is implicitly factorizing a form of Laplacian matrix \cite{qiu2018network}. The interested readers can refer to the surveys of heterogeneous graph models \cite{wang2020survey,yang2020heterogeneous}. 

\subsubsection{Graph Convolutional Networks for Recommendation:} In recent years, we have seen the superior performance of graph convolutional networks \cite{kipf2016semi, hamilton2017inductive, velickovic2018graph} on several machine learning tasks on graphs. Relational GCN \cite{schlichtkrull2018modeling} extends GCN to graph with different edge types by proposing a learnable weight matrix for each relation to improve link prediction performance. Neural Graph Collaborative Filtering (NGCF) \cite{wang2019neural} operates on user-item bipartite graph and  explicitly incorporates the high-order connectivity between users and items to learn their embeddings. LightGCN \cite{he2020lightgcn} simplifies GCN for recommendation by proposing linear embedding propagation between user and items. 

\subsubsection{Heterogeneous Graph Neural Networks:} HAN \cite{wang2019heterogeneous} proposed a hierarchical, node-level and semantic-level attention based heterogeneous graph neural network. HAN requires performing random-walks based on meta-paths. HetGNN \cite{zhang2019heterogeneous} proposed a random-walk with restart strategy that first samples heterogeneous neighbor nodes and then aggregates them based on neighbor features with Bi-LSTM. GATNE\cite{Cen:2019:RLA:3292500.3330964} utilizes heterogeneous skip-gram objective function to learn node embeddings of attributed multiplex heterogeneous graphs. The training samples for skip-gram are curated by performing random-walks on nodes with same relations.
HGCN \cite{zhu2020hgcn} proposed a heterogeneous graph convolutional network that can learn fine-grained relational features of the heterogeneous graph. 
Heterogeneous Graph Transformer (HGT) \cite{hu2020heterogeneous} proposed a scalable model that can handle dynamic heterogeneous graphs. HGT introduced heterogeneous mutual attention and heterogeneous message passing to perform convolutions on heterogeneous graphs. Note that, these heterogeneous graph models are trained on single-machine system and training these models on a distributed setup requires solving many challenges which are discussed in the next section. 

\subsubsection{Distributed Frameworks for Heterogeneous Graph Models:}
The distributed training frameworks for heterogeneous graphs in general addresses three main scalability challenges related to storage, sampling and GNN operations. The AliGraph \cite{zhang2019heterogeneous} system address these challenges by proposing three layers: i) Storage layer partitions the graphs using four graph partition algorithms and introduces indexing and caching operations for retrieval of node and its neighbors features, ii) Sampling layer identify the node's neighbors which are required by GNN. Sampling is efficient as each worker sample neighbors on the assigned graph partition, and iii) Operations layers that implement aggregation and combination operation of GNN in an optimized manner. DistDGL \cite{zheng2020distdgl} stores the graphs on multiple devices by partitioning the graph using METIS \cite{karypis1998fast} algorithm. The node features are stored using a distributed key-value store. For sampling, DistDGL introduced a distributed sampler that follows client-server model to samples the node neighbors in parallel. DistDGL perform random-walks on each partition and a single random-walk does not move across different partitions. We estimate that the migration to these new frameworks in the Pinterest software ecosystem would also incur significant infrastructure costs.  P3 \cite{gandhi2021p3} proposed a novel distributed training approach that eliminates partitioning overheads and high communication among workers. P3 introduced a pipelined push-pull parallelism based execution strategy for the distributed training of GNNs.  

%% file: methodology.tex
\begin{table*}\sffamily
\centering
\begin{tabular}{l|l|*5{C}@{}}
\toprule
Candidate         & Graph & Rank-1 & Rank-2 & Rank-3 & Rank-4  & Rank-5 \\ 
\midrule
 &  Pin-Board     &  \includegraphics[width=0.7in]{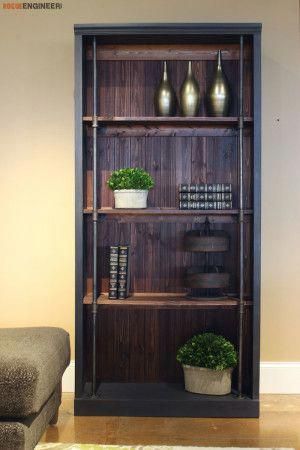}      &  \includegraphics[width=0.7in]{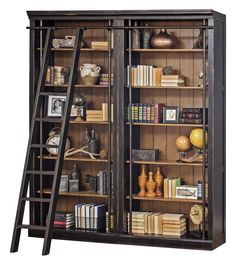}         &  \includegraphics[width=0.7in]{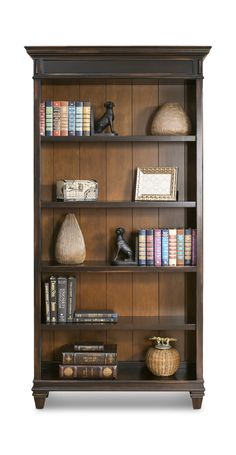}         &     \includegraphics[width=0.7in]{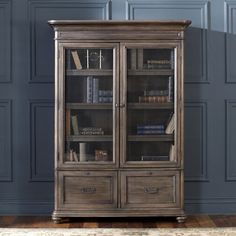}         &     \includegraphics[width=0.7in]{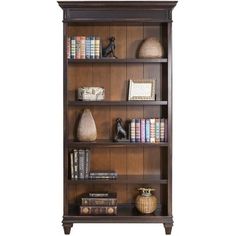}      \\ 
\cline{2-7} \\ 
   \multirow{4}{*}{\includegraphics[width=1.5in]{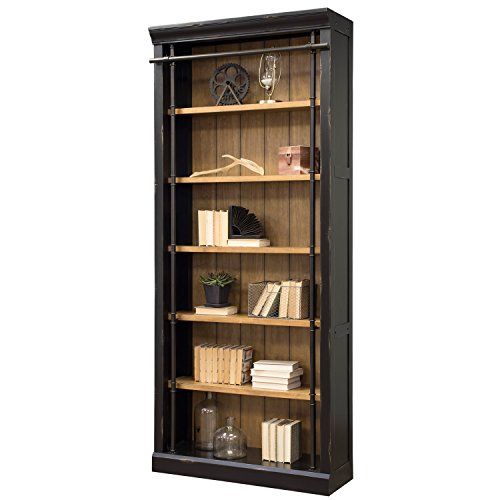}}               &   User-Product    &   \includegraphics[width=0.7in]{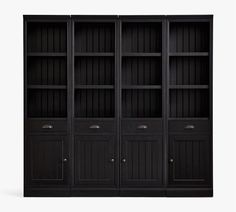}      &     \includegraphics[width=0.7in]{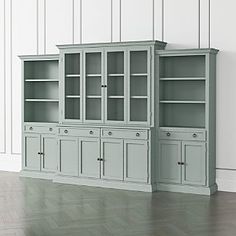}   & \includegraphics[width=0.7in]{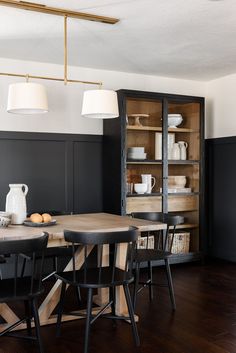}       &     \includegraphics[width=0.7in]{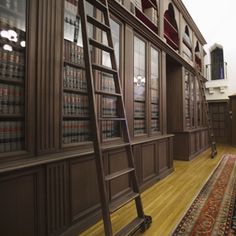}   &     \includegraphics[width=0.7in]{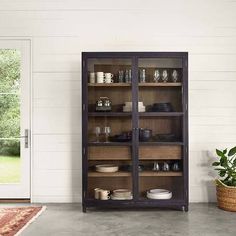}    \\ 
\cline{2-7}\\ 
                  &   User-Ad    &  \includegraphics[width=0.7in]{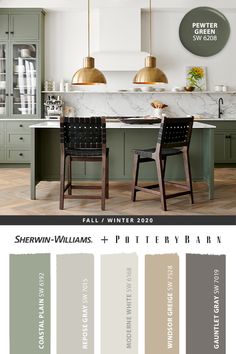}      &     \includegraphics[width=0.7in]{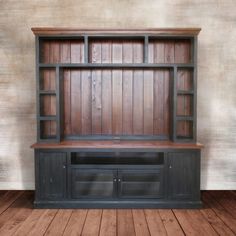}   & \includegraphics[width=0.7in]{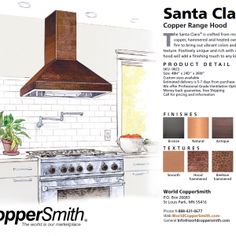}       &     \includegraphics[width=0.7in]{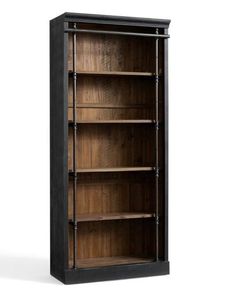}    &     \includegraphics[width=0.7in]{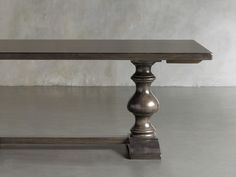}  \\ 
\cline{2-7}\\ 
                  &   SearchQuery-Pin    &     \includegraphics[width=0.7in]{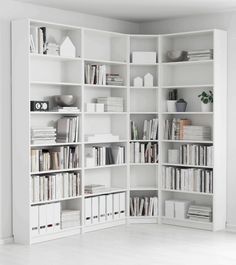}      &     \includegraphics[width=0.7in]{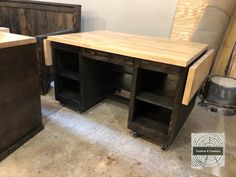}   & \includegraphics[width=0.7in]{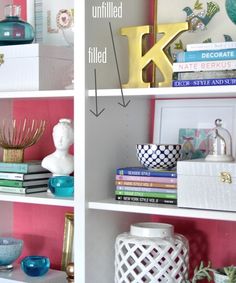}       &     \includegraphics[width=0.7in]{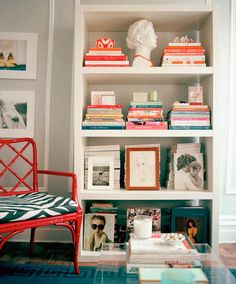}    &     \includegraphics[width=0.7in]{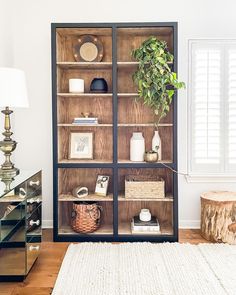}    \\ 
\bottomrule
\end{tabular}
\caption{Candidate pin and its neighbors from diverse bipartite graphs.}
\label{tab:neighbrs}
\end{table*}

\alglanguage{pseudocode}
\begin{algorithm}[t]
    \small
    \caption{Training Pipeline of MultiBiSage}
    \label{algo:train_pipeline}
\begin{algorithmic}[1]
\State Construct the bipartite graphs from the  logs at Pinterest.
\State Get the local graph structure of the nodes from the bipartite graphs.
\vspace{0.1em}
\begin{algsubstates}
\State Perform random-walks on each bipartite graph with the help of in-house random-walker Pixie system.
\State Select top-k highly visited neighbors of the nodes from the random-walks.
\end{algsubstates}
\State Collect training data. 
\begin{algsubstates}
\State Curate query-pin and engaged-pin from multiple surfaces at Pinterest. \State Collect the visual and textual features of query pin, engaged pin and their neighbors.
\State Collect a random sample of million pins and treat these pins as negative samples.
\end{algsubstates}
\vspace{0.1em}
\State Train MultiBiSage model with a modified Sampled Softmax \cite{jean2014using}.
\end{algorithmic}
    
\end{algorithm}
\begin{figure*}[t]
    \centering
    \includegraphics[width=0.85\linewidth]{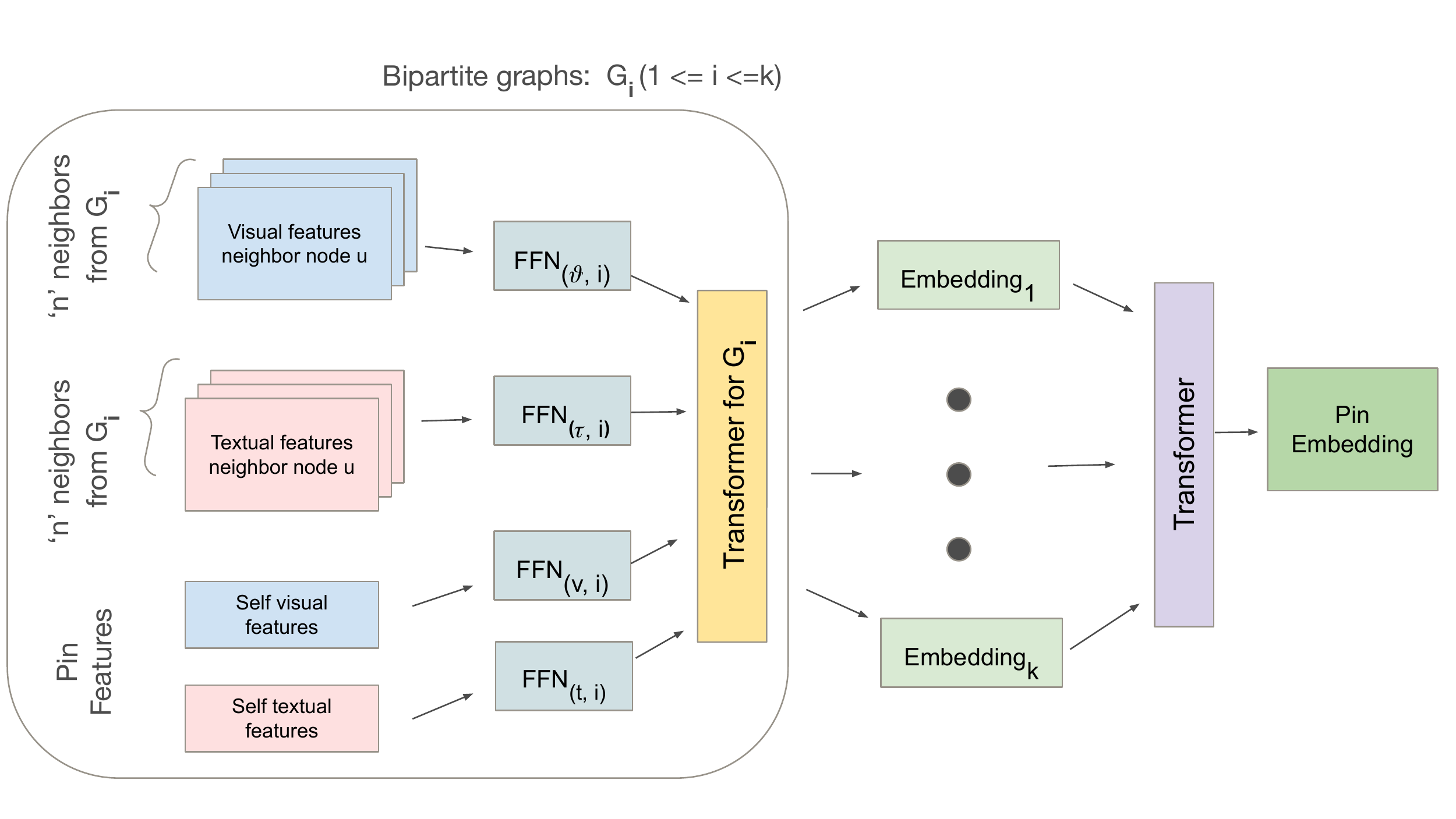}
    \caption{The model architecture of MultiBiSage.}
    \label{fig:model}
\end{figure*}

Given the scale of Pinterest data, we require an optimized data curation and ingestion pipeline to feed the bipartite graph features to the machine learning model. Moreover, we also require a novel model architecture to better capture the bipartite graph features in order to learn high-quality pin embeddings. We describe these details in this section.

\subsection{Training Pipeline}

Algorithm \ref{algo:train_pipeline} shows the training pipeline of MultiBiSage. The first step is the construction of bipartite graphs from the user's interaction logs at Pinterest. However constructing these graphs on a single machine is time-consuming and not practical. Hence, we develop a spark-based graph generator, which given the two entity-types and the interaction type, curates the corresponding bipartite graph by parsing the logs. The spark-based graph generator also includes a degree-based graph pruning algorithm that can reduce the number of the edges in the graph. The algorithm can be described as follows: let $a$ and $b$ be the specified minimum and maximum node degrees and let $p$ be the pruning factor between 0 and 1. The pruning algorithm selects node (say $u$) with node degree greater than $a$ and then randomly removes the  edges incident on node $u$ such that new node degree of node $u$ is $min(a * p, b)$.

Once the bipartite graphs are created, our next goal is to identify the neighbors of pin nodes from the bipartite graphs. As mentioned earlier, the "on-the-fly" sampling of neighbors during  model training from billion-nodes and billion-edges graph is extremely inefficient and not practical. Hence, we identify the node neighbors before training MultiBiSage using Pixie \cite{eksombatchai2018pixie} system. Pixie system loads the entire graph in 2 TB RAM machine and performs random-walks using optimized C++ threads. For each pin node, we select top-k highly visited pin neighbors. The pseudo-code of Pixie's random-walker is shared in the Appendix \ref{algo:pixie}.  Table \ref{tab:neighbrs} shows a sample of the top-5 highly visited pins of a candidate pin.

Next, we collect the training samples and the features of the training samples with the help of spark jobs. The training sample is constructed as follows. A user interacts with a pin (referred to as query pin) and is shown recommendations, if the user interacts with one of the recommended pin (referred to as engaged pin)  we treat the query and engaged pin as positive pair. The goal of MultiBiSage model is to learn pin embeddings such that distance between query and engaged pin is close to each other. We describe the model architecture next.

\subsection{MultiBiSage model architecture}
Figure \ref{fig:model} shows the model architecture of MultiBiSage. Let $p$ and $x_p$ be the pin and embedding of pin $p$, respectively. Here $x_p \in \mathbb{R}^d$ is $l_2$ normalized $d$-dimensional vector. Let  $v_p \in \mathbb{R}^{d_v}$ and $t_p \in \mathbb{R}^{d_t}$ be the visual and textual features of pin $p$, respectively. Assume there are $k$ bipartite graphs with $G_i$ being the $i^{th}$ bipartite graph. Let $n$ be the number of neighbors from each bipartite graph. Let $\vartheta_{i,p} \in \mathbb{R}^{n \times d_v}$ and $\tau_{i,p} \in \mathbb{R}^{n \times d_t}$ be the visual and textual features of the neighbors of pin $p$ from $G_i$, respectively. 

Mathematically, the MultiBiSage model can be described as follows. First, we pass the visual and textual features of pin and its neighbors through a feedforward neural network (FFN) to learn the intermediate representation of dimension size $d_h$. 
\begin{equation}
\begin{split}
x_{p,v_i} &= FFN(v_p) = ReLU(v_p W_{v_i} + b_{v_i}) \\
x_{p,t_i} &= FFN(t_p) =ReLU(t_p W_{t_i} + b_{t_i}) \\
x_{p,\vartheta_{i}} &= FFN(\vartheta_{i,p}) =ReLU(\vartheta_{i,p} W_{\vartheta_i} + b_{\vartheta_i}) \\
x_{p,\tau_i}&= FFN(\tau_{i,p})=ReLU(\tau_{i,p} W_{\tau_i} + b_{\tau_i}) \\
\end{split}
\end{equation}

where $W_{v_i} \in \mathbb{R}^{d_v \times d_h}$, $W_{t_i} \in \mathbb{R}^{d_t \times d_h}$, $W_{\vartheta_i} \in \mathbb{R}^{d_v \times d_h}$, and $W_{\tau_i}  \in \mathbb{R}^{d_t \times d_h}$ be the learnable weight matrices. Here, $b_{v_i} \in \mathbb{R}^{d_h}, b_{t_i}\in \mathbb{R}^{d_h},  b_{\vartheta_i} \in \mathbb{R}^{d_h}$ and $b_{\tau_i} \in \mathbb{R}^{d_h}  $ are the biases parameters. 

The intermediate representations $x_{v,i}, x_{t,i}, x_{\vartheta_i}, x_{\tau_i}$ and a global token $x_{g_i} \in \mathbb{R}^{d_h}$ are then concatenated and passed to the transformer layer.
\begin{equation}
s_{p,i} = Concat(x_{g_i}, x_{v,i}, x_{t,i}, x_{\vartheta_i},x_{\tau_i})
\end{equation}

where $s_{p,i} \in \mathbb{R}^{(1+ 2 \times (1+n)) \times d_h}$. Here, $s_{p,i}$ can be considered as a sequence with $1+ 2 \times (1+n)$ number of tokens and each token is represented in $d_h$ dimensional space.

We pass the sequence $s_{p,i}$ through the multihead attention layer \cite{vaswani2017attention} where each input token has three intermediate representations referred to as query (Q), key (K), and value (V). The attention (A) between all tokens is computed using the below equation. 
\begin{equation}
A(Q, K, V) = softmax (\frac{QK^T}{\sqrt{d_k}})V \\
\end{equation}
where $d_k$ is dimension size of queries and keys. We apply attention with $H$ number of heads as shown below:
\begin{equation}
\begin{split}
MultiHead_i(Q, K, V) &= Concat (head_1, ..., head_H)W^O \\
head_j & = A(QW_j^Q,KW_j^K,VW_j^V)
\end{split}
\end{equation}

where $W_j^Q \in \mathbb{R}^{d_h \times d_k}, W_j^K \in \mathbb{R}^{d_h \times d_k}, W_j^V \in \mathbb{R}^{d_h \times d_v}$ and $W^O \in \mathbb{R}^{Hd_v \times d}$ are trainable parameters and $d_k=d_v=d_h/H$. 

The representation of global token $x_{g_i}$ after passing it through multihead attention is treated as pin embedding $x_{p,i} \in \mathbb{R}^d$ of pin $p$ from bipartite graph $G_i$. The  pin embeddings $x_{p,i}$ are then passed to second transformer along with global token  $x_g \in \mathbb{R}^d$ as shown below:
\begin{equation}
    s = Concat(x_g, x_{p,1}, x_{p,2}, ..., x_{p,k})
\end{equation}
where $s \in \mathbb{R}^{(1+k)*d}$. The representation of global token $x_{g}$ after passing it through transformer layer is $L_2$ normalized and considered as pin embedding $x_p \in \mathbb{R}^d$.

\subsubsection{Training objective}
Let $q_i$, and $e_i$ be the $i^{th}$ query pin and engaged pin in a batch $\mathcal{B}$, respectively. Let $x_{q_i} \in \mathbb{R}^d$ and $x_{e_i} \in \mathbb{R}^d$ be the embedding of these pins computed by MultiBiSage. Also, let $C_\mathcal{B}=\{e_1, ... e_{|\mathcal{B}|}\}$ be the set of all the engaged pins in the batch. Then,  we minimize the sampled softmax loss function \cite{jean2014using} as shown below:

\begin{equation}
        - \frac{1}{|\mathcal{B}|} \sum_{i=1}^{|\mathcal{B}|} log \frac{e^{\langle x_{q_i}, x_{e_i}\rangle \;- \;log\; Q_p(e_i|q_i)}}{ \sum_{e' \in C_\mathcal{B}} e^{\langle x_{q_i}, x_{e'}\rangle\;- \;log\; Q_p(e'|q_i)}}
        \label{eqn:loss1}
\end{equation}

where $\langle .\;, . \rangle$ represents the dot product between learned embeddings. Here, we are performing softmax over sampled classes $C_\mathcal{B}$ instead of performing softmax over all the classes. The softmax over-sampled classes introduce sampling bias in the full softmax computation. This bias is corrected through the proposal distribution $Q_p(e_i|q_i)$ which is the probability of $e_i$ being included as a positive sample in the training batch. We utilize count-min sketch \cite{cormode2005improved} to estimate $Q_p(e_i|q_i)$ in a streaming manner.

Notice that, for query-pin $q_i$, we are treating all the other engaged pins $e_j$ ($e_j\ne e_i$) in the batch as negative samples. If a pin is popular, it would frequently appear as an engaged pin (say $e_k$) in the batch. The above loss function would then unfairly penalize engaged pin $e_k$, as they are more likely to be selected as negative pin during training. To address this issue,  we adopt the mixed negative sampling approach \cite{yang2020mixed}, in which we select a  set of random negatives $\mathcal{M}$ where $|\mathcal{M}|=|\mathcal{B}|$ and compute the below loss function.

\begin{equation}
        - \frac{1}{|\mathcal{M}|} \sum_{i=1}^{|\mathcal{M   }|} log \frac{e^{\langle x_{q_i}, x_{e_i}\rangle \;- \;log\; Q_n(e_i    )}}{ \sum_{e' \in \mathcal{M}} e^{\langle x_{q_i}, x_{e'}\rangle\;- \;log\; Q_n(e')}}
        \label{eqn:loss2}
\end{equation}
where $Q_n(e)$ represents the probability of sampling pin $e$. We also utilize count-min sketch \cite{cormode2005improved} to estimate $Q_n(e)$ in a streaming manner. We optimize  both the loss function in Equation \ref{eqn:loss1} and Equation \ref{eqn:loss2} by summing them with equal weightage.       

%% file: experiments.tex
\subsection{Experiment Details}

\subsubsection{Dataset}
We perform experiments on the six bipartite graphs curated from the logs present at Pinterest. The statistics of these graphs are present in Table \ref{tab:pin_stat}. All the graphs are collected over the period of one year and represent a random subset of users, pins, boards, search-queries, ads, products, creators, idea pins, and video pins. The edge type LC refers to the long click. A long click refers to the interaction where a user clicks on a pin and does not return to the Pinterest website in 10 secs. LC often acts as a signal that the user found the item that he/she was searching for. The raw Pin-Board graph consists of 100+ billion edges. We perform the degree-based pruning to reduce the number of edges from 100+ billion to 7 billion by setting the parameters as follows: minimum node degree=10, maximum node degree=10000 and prune factor=0.86.  We apply the degree-based pruning to only pin-board graph. In the case of the SearchQuery-Pin-LC graph, each search-query text is considered as a node, and no text processing is performed on the search-query text. We select 50 neighbors from every bipartite graph. Table \ref{tab:train_data} contains the training data statistics on which we train all the models. The training data is  collected over a certain temporal period.

\begin{table}[t]
    \centering
    \resizebox{1.0\linewidth}{!}{
    \begin{tabular}{l|r|r|r}
    \toprule
    Graphs & Num. Entity 1 & Num. Entity 2 &  Num. Edges   \\
    \midrule
    Pin-Board-ContainedIn & 1,995,423,350 & 1,927,181,284  & 7,144,399,779 \\
    User-Product-LC & 74,130,469 & 40,198,339  & 472,307,057 \\
    User-AD-LC & 114,059,486 & 9,392,321 & 710,736,288 \\
    SearchQuery-Pin-LC & 397,290,595 & 189,846,404 & 1,307,227,847 \\
    Creator-IdeaPin-Created & 1,508,535 & 12,349,297 & 13,984,765 \\
    IdeaPin-User-FollowedBy & 2,502,566  & 41,002,754 & 98,570,789 \\
    \bottomrule
    \end{tabular}
    }
    \caption{The statistics of Pinterest's bipartite graphs.  Graph nomenclature convention: Entity 1 - Entity 2 - EdgeType.  The graphs represent a random subset of users, pins, boards, search-queries, ads, products, creators, idea pins, and video pins.}
    \label{tab:pin_stat}
    \vspace{-1em}
\end{table}

\begin{table}[t]
    \centering
    \resizebox{0.9\linewidth}{!}{
    \begin{tabular}{l|r}
    \toprule
    Statistic & Count\\
    \midrule
    Number of distinct Query Pins & 158,489,849\\ 
    Number of distinct Engaged Pins   & 150,706,623 \\
    Number of distinct Query-Engaged Pins & 573,056,337 \\
    \bottomrule
    \end{tabular}
    }
    \caption{The training data volume collected over a certain temporal period.  }
    \label{tab:train_data}
    \vspace{-1em}
\end{table}
\subsubsection{Experimental Setting:} We train the models with Adam optimizer with a learning rate of 0.002 and batch size of 8032. We gradually increase learning rate in optimizer \cite{goyal2017accurate} using Cosine Annealing \cite{loshchilov2016sgdr}.

\subsubsection{Baselines} We primarily focus on the comparison of the state-of-the-art production model PinSage currently deployed at Pinterest. We also include advanced deep learning models and a few of their variants. The baselines we consider are
\begin{itemize}
    \item \textbf{Transformer} \cite{vaswani2017attention}: We pass the visual and textual embeddings of the neighbors from the bipartite graphs to the model. The five additional bipartite graphs increase the sequence length passed to the transformer by 5x. This model can be considered as an extension of the deployed PinSage model to the six bipartite graphs.
    \item \textbf{SharedTransformer}: Instead of having a transformer for every bipartite graph, we have one transformer model that is shared across multiple bipartite graph features.
    \item \textbf{NFFNTransformer}: Here, the bipartite graph embeddings are passed through a feed forward neural network layer which aggregates the embeddings from all the bipartite graphs.
    \item \textbf{NSumTransformer}: We perform an element-wise sum of all the bipartite graph embeddings and then pass the resultant embeddings to the transformer.
    \item \textbf{NHadamardTransformer}: We perform the Hadamard product of all the bipartite graph embeddings and then pass the resultant embeddings to the transformer.
    \item \textbf{PinFeatToLastTransformer}: Here, we modify the MultiBiSage model and pass the pin-features to only the last transformer layer instead of repeatedly passing them to every bipartite graph transformer.
    \item \textbf{AggregateByFFN}: Here, we modify the MultiBiSage model and replace the last transformer layer with a feed forward neural network that aggregates the bipartite graph pin embeddings ($x_{p,i}$). 
\end{itemize}

\begin{table}[t]
    \centering
    \resizebox{1.0\linewidth}{!}{
    \begin{tabular}{l|c|c|c}
    \toprule
    Surface/Entity Type & Engagement Type & PinSage & MultiBiSage   \\
    \midrule
    \multirow{2}{*}{Organic Surface} & Add-to-cart     &  0.893   &  0.907 (+1.57\%) \\
     & Checkout     &   0.893    & 0.907 (+1.57\%)\\
    \midrule
    Ads &  Good-click-through  &  0.651   & 0.684	(+5.07\%) \\
    \midrule
    Related Products  & Long-click &  0.733  & 0.749	(+2.18\%) \\
    
    \midrule
    \multirow{2}{*}{Idea Pin} & Close-ups &  0.724    & 0.776	(+7.18\%) \\
    & Repin &    0.797  & 0.849	(+6.52\%) \\
    \midrule
    \multirow{2}{*}{Video Pin} & Close-ups & 0.446     & 0.47	(+5.38\%) \\
    & Repin & 0.571     & 0.603	(+5.60\%) \\
    \bottomrule
    \end{tabular}
    }
    \caption{Recommendation performance: Recall@10}
    \label{tab:compare_w_head}
    \vspace{-1em}
\end{table}

\subsubsection{Metrics}
We measure the performance of our recommendations on multiple types of engagement. These engagements happen on different surfaces at Pinterest. The organic surface refers to pins present on the home feed (the landing page with an endless set of recommended pins) surface.  The related products surface corresponds to the page where a user clicks on a pin and is shown pin recommendations.  The entities such as Ads, Idea Pins, and Video Pins can be present on both organic and related products surface. The users can have various types of engagement over these surfaces and entities. The   Add-to-Cart and Checkout engagement corresponds to the user's actions related to purchasing. The Good-click-through engagement occurs when the user is displayed an ad and after clicking the ad, the user does not return to the website in less than 30 secs.  In Long click engagement, the user clicks on a pin and does not return back to the website in less than 10 secs. Close-ups refer to the action where the user zooms in into the pin. Repin refers to the action where the user saves the pin to his/her board. The users often perform these engagements on our set of recommended pins.

\begin{table*}[t]
    \centering
    \resizebox{1.0\linewidth}{!}{
    \begin{tabular}{l|c|c|c|c|c|c|c|c}
    \toprule
    Surface/Entity Type & Engagement Type  & Transformer & Shared & NSum & NHadamard & NFFN & PinFeatToLast & Aggregate  \\
     &  &  & Transformer & Transformer & Transformer & Transformer & Transformer & ByFFN    \\
    \midrule
    
   \multirow{2}{*}{Organic Surface} & Add-to-cart	&	0.900	 (+0.8\%) &	0.902	 (+1.0\%)  &	0.901	 (+0.9\%)  &	0.899	 (+0.7\%)  &	0.900	 (+0.8\%)  &	0.874	(-2.1\%)  &	0.885	(-0.9\%)  	\\
& Checkout   	&	0.898	 (+0.6\%)  &	0.900	 (+0.8\%)  &	0.900	 (+0.8\%)  &	0.901	 (+0.9\%)  &	0.902	 (+1.0\%)  &	0.875	(-2.0\%)  &	0.885	(-0.9\%)  	\\
 \midrule																														
Ads  & Good-click-through	&	0.670	 (+2.9\%)  &	0.676	 (+3.8\%)  &	0.669	 (+2.8\%)  &	0.668	 (+2.6\%)  &	0.672	 (+3.2\%)  &	0.629	(-3.4\%)  &	0.661	(+1.5\%)  	\\
 \midrule																														
Related Products & Long-click 	&	0.740	 (+1.0\%)  &	0.738	 (+0.7\%)  &	0.738	 (+0.7\%)  &	0.737	 (+0.5\%)  &	0.741	 (+1.1\%)  &	0.708	(-3.4\%)  &	0.741	(+1.1\%)  	\\
 \midrule																														
\multirow{2}{*}{Idea Pin} & Close-ups 	&	0.749	 (+3.5\%)  &	0.757	 (+4.6\%)  &	0.747	 (+3.2\%)  &	0.746	 (+3.0\%)  &	0.752	 (+3.9\%)  &	0.729	(+0.7\%)  &	0.760	(+5.0\%)  	\\
& Repin 	&	0.827	 (+3.8\%)  &	0.833	 (+4.5\%)  &	0.829	 (+4.0\%)  &	0.824	 (+3.4\%)  &	0.827	 (+3.8\%)  &	0.808	(+1.4\%)  &	0.837	(+5.0\%)  	\\
 \midrule																														
\multirow{2}{*}{Video} & Close-ups 	&	0.449	 (+0.7\%)  &	0.454	 (+1.8\%)  &	0.451	 (+1.1\%)  &	0.453	 (+1.6\%)  &	0.451	 (+1.1\%)  &	0.432	(-3.1\%)  &	0.463	(+3.8\%)  	\\
& Repin 	&	0.578	 (+1.2\%)  &	0.581	 (+1.8\%)  &	0.575	 (+0.7\%)  &	0.573	 (+0.4\%)  &	0.578	 (+1.2\%)  &	0.553	(-3.2\%)  &	0.591	(+3.5\%)  	\\
    \bottomrule
    \end{tabular}
    }
    \caption{Baselines. The percentages refer to the percentage improvement over deployed PinSage model.}
    \label{tab:baselines}
\end{table*}

    

\begin{table}[t]
    \centering
    \resizebox{1.0\linewidth}{!}{
    \begin{tabular}{l|c|c|c|c}
    \toprule
Surface/Entity Type & Engagement Type & 10  & 20 & 50 \\
    \midrule
   \multirow{2}{*}{Organic} & Add-to-cart     &  0.899	&	0.903	& 0.907	\\
& Checkout     &  0.900	&	0.904	& 0.907	\\
\midrule
Ads  & Good-click-through &  0.681	&	0.68	& 0.684	\\
\midrule
Related Products & Long-click &  0.738	&	0.743	& 0.749	\\
\midrule
\multirow{2}{*}{Idea Pin} & Close-ups &  0.771	&	0.772	& 0.776	\\
& Repin &    0.842	&	0.846	& 0.849	\\
\midrule
\multirow{2}{*}{Video} & Close-ups & 0.469	&	0.468	& 0.470	\\
& Repin & 0.593	&	0.597	& 0.603	\\
    \bottomrule
    \end{tabular}
    }
    \caption{Impact of the number of selected neighbors.}
    \label{tab:num_neigh}
\end{table}

Our goal is to improve the quality of user engagement by providing better recommendations. Here, we measure the quality of the recommendations using the recall@10 metric. Given a pair of query pin  $q$ and engaged pin $e$, we first generate the query pin embedding $x_{q}$ and engaged pin embedding $x_{e}$. We additionally generate the embeddings of one million randomly sampled pins. We then find the ten nearest neighbors pins of query pin $q$ using $x_{q}$ in the embedding space. The metric Recall@10 computes the fraction of times the ground-truth engaged pin $e$ appears in the top-10 nearest neighbors of query pin.

\subsection{Results}

\begin{table*}[t]
    \centering
    \resizebox{1.0\linewidth}{!}{
    \begin{tabular}{l|c|c|c|c|c|c}
    \toprule
   Surface/Entity Type &  Engagement Type & MultiBiSage+  & MultiBiSage+  & MultiBiSage+ & MultiBiSage+ & MultiBiSage+   \\
    &   & UserProd+PB  & UserAd+PB  & QueryPin+PB & CCI+PB & IUF+PB    \\
    \midrule
    \multirow{2}{*}{Organic Surface} & Add-to-cart	&	0.897	 (+0.45\%)  &	0.899	 (+0.67\%)  &	0.899	 (+0.67\%)  &	0.897	 (+0.45\%)  &		0.897	 (+0.45\%)   	\\
& Checkout   	&	0.898	 (+0.56\%)  &	0.896	 (+0.34\%)  &	0.897	 (+0.45\%)  &	0.897	 (+0.45\%)  &		0.897	 (+0.45\%)   \\
\midrule
Ads  & Good-click-through	&	0.660	 (+1.38\%)  &	0.679	 (+4.30\%)  &	0.677	 (+3.99\%)  &	0.662	 (+1.69\%)  &		0.658	 (+1.08\%)  \\
\midrule
Related Products & Long-click 	&	0.737	 (+0.55\%)  &	0.739	 (+0.82\%)  &	0.737	 (+0.55\%)  &	0.738	 (+0.68\%)  &		0.735	 (+0.27\%)  \\
\midrule
\multirow{2}{*}{Idea Pin} & Close-ups 	&	0.742	 (+2.49\%)  &	0.751	 (+3.73\%)  &	0.746	 (+3.04\%)  &	0.747	 (+3.18\%)  &		0.756	 (+4.42\%)  \\
& Repin 	&	0.814	 (+2.13\%)  &	0.822	 (+3.14\%)  &	0.815	 (+2.26\%)  &	0.824	 (+3.39\%)  &		0.833	 (+4.52\%)   \\
\midrule
\multirow{2}{*}{Video} & Close-ups 	&	0.459	 (+2.91\%)  &	0.463	 (+3.81\%)  &	0.461	 (+3.36\%)  &	0.465	 (+4.26\%)  &		0.461	 (+3.36\%)  \\
& Repin 	&	0.586	 (+2.63\%)  &	0.591	 (+3.50\%)  &	0.588	 (+2.98\%)  &	0.587	 (+2.80\%)  &		0.589	 (+3.15\%)  \\
\bottomrule
    \end{tabular}
    }
    \caption{Ablation Study on Bipartite Graphs. PB, UserProd, UserAd, QueryPin, CCI, and IUF refers to Pin-Board-Contained, User-Product-LC, User-Ad-LC, SearchQuery-Pin-LC, Creator-IdeaPin-Created, IdeaPin-User-Follow bipartite graphs, respectively.} 
    \label{tab:ablation_with_PB}
\end{table*}

\subsubsection{Comparison with Baselines:}
In Table \ref{tab:compare_w_head}, we compare our proposed model MultiBiSage against the currently deployed PinSage version at Pinterest. The percentages in Table \ref{tab:compare_w_head} in column MultiBiSage refers to the percentage improvement over PinSage model.  We observe that MultiBiSage outperforms PinSage on all the performance metrics. The difference in performance between MultiBiSage and PinSage is statistically significant with standard paired t-test at significance level 0.01. On Ads, Idea Pins, and Video Pins, we observe more than 5\% improvement over PinSage. These results shows that training MultiBiSage on graphs that captures diverse interactions result in learning higher-quality pin embeddings than training PinSage on only Pin-Board graph.


\begin{table*}[t]
    \centering
    \resizebox{1.0\linewidth}{!}{
    \begin{tabular}{l|c|c|c|c|c|c|c}
    \toprule
   Surface/Entity Type &  Engagement Type    & MultiBiSage+  & MultiBiSage+  & MultiBiSage+ & MultiBiSage+ & MultiBiSage+  & MultiBiSage+   \\
    &   & UserProd  & UserAd  & QueryPin & CCI & IUF & UserProd+ UserAd+ \\
    &   &   &   &  &  &  & QueryPin +CCI+IUF  \\
    \midrule
    \multirow{2}{*}{Organic Surface} & Add-to-cart	&	0.810	(-9.29\%)  &	0.822	(-7.95\%)  &	0.845	(-5.38\%)  &	0.807	(-9.63\%)  &	0.811	(-9.18\%)  &	0.874	(-2.13\%)  	\\
& Checkout   	&	0.811	(-9.18\%)  &	0.819	(-8.29\%)  &	0.844	(-5.49\%)  &	0.807	(-9.63\%)  &	0.814	(-8.85\%)  &	0.874	(-2.13\%)  	\\
 \midrule																										
Ads  & Good-click-through	&	0.606	(-6.91\%)  &	0.630	(-3.23\%)  &	0.627	(-3.69\%)  &	0.605	(-7.07\%)  &	0.609	(-6.45\%)  &	0.670	 (+2.92\%)  	\\
 \midrule																										
Related Products & Long-click 	&	0.666	(-9.14\%)  &	0.665	(-9.28\%)  &	0.685	(-6.55\%)  &	0.660	(-9.96\%)  &	0.663	(-9.55\%)  &	0.711	(-3.00\%)  	\\
 \midrule																										
\multirow{2}{*}{Idea Pin} & Close-ups 	&	0.637	(-12.02\%)  &	0.637	(-12.02\%)  &	0.648	(-10.50\%)  &	0.654	(-9.67\%)  &	0.690	(-4.70\%)  &	0.748	 (+3.31\%)  	\\
& Repin 	&	0.722	(-9.41\%)  &	0.723	(-9.28\%)  &	0.733	(-8.03\%)  &	0.738	(-7.40\%)  &	0.780	(-2.13\%)  &	0.828	 (+3.89\%)  	\\
 \midrule																										
\multirow{2}{*}{Video} & Close-ups 	&	0.397	(-10.99\%)  &	0.407	(-8.74\%)  &	0.404	(-9.42\%)  &	0.404	(-9.42\%)  &	0.406	(-8.97\%)  &	0.450	 (+0.90\%)  	\\
& Repin 	&	0.511	(-10.51\%)  &	0.510	(-10.68\%)  &	0.521	(-8.76\%)  &	0.511	(-10.51\%)  &	0.516	(-9.63\%)  &	0.563	(-1.40\%)  	\\
\bottomrule
    \end{tabular}
    }
    \caption{Ablation Study on Bipartite Graphs.  UserProd, UserAd, QueryPin, CCI and IUF refers to User-Product-LC, User-Ad-LC, SearchQuery-Pin-LC, Creator-IdeaPin-Created, and IdeaPin-User-Follow bipartite graphs, respectively.   } 
    \label{tab:ablation}
\end{table*}

We compare the performance of MultiBiSage with the baselines in Table \ref{tab:baselines}. The percentages shown in parenthesis refer to the percentage improvement over the deployed PinSage model. The difference in performance between MultiBiSage and the baselines is statistically significant with standard paired t-test at significance level 0.01. We also observe that most of the baselines result in the improvement of recommendation performance over PinSage. This is partly due to the fact that these models can utilize the additional neighborhood context of pins from diverse bipartite graphs. Next, we observe that the distribution of the node's neighbors from each bipartite graph is different. This can be inferred from the performance of the Shared-Transformer model where the transformer is shared across multiple bipartite graph features. The models NSum-Transformer, NHadamard-Transformer, and NFFN-Transformer recommendation is similar to that of Transformer. The difference between them is not statistically significant with standard paired t-test at significance level 0.01. The MultiBiSage model variants PinFeatToLastTransformer and AggregateByFFN recommendation performance is poorer than that of MultiBiSage.

\subsubsection{Impact of the number of neighbors:} We study the impact of the number of neighbors of nodes on the MultiBiSage performance. If the lower number of neighbors from different bipartite graphs can achieve similar performance, then the model inference would be faster. Table \ref{tab:num_neigh} shows the model performance on 10, 20 and 50 number of neighbors. We observe that the recommendation performance of MultiBiSage with 10 and 20 number of neighbors is similar. The difference in performance between them is not statistically significant with paired t-test at significance level 0.01. However, the difference in the performance of MultiBiSage with 50 neighbors is significantly better than that of MultiBiSage with 10 and 20 neighbors. 

\subsubsection{Ablation Study:} Next, we study the impact of incorporating each bipartite graph on the MultiBiSage performance. The result of this ablation study is in shown in Table \ref{tab:ablation_with_PB} and Table \ref{tab:ablation}. In Table \ref{tab:ablation_with_PB}, we train the MultiBiSage model with each introduced bipartite graph along with Pin-Board graph, while in Table \ref{tab:ablation} we train MultiBiSage with only the introduced bipartite graphs. We also have an experiment where we train MultiBiSage on all the five bipartite graphs except the Pin-Board graph in  Table \ref{tab:ablation}.  From Table \ref{tab:ablation_with_PB}, we observe MultiBiSage+UserAd+PB achieves over 4.3\% improvement  on Ads over deployed PinSage model. At the same time, both MultiBiSage+CCI+PB and MultiBiSage+IUF+PB that are trained on graphs containing Idea pins show from 3.18\% to 4.42\% improvement on Idea pins over the PinSage model. Since Idea pins can also contain videos, we observe that these two models show improvement over videos pin recommendation tasks. The model MultiBiSage+UserProd+PB and MultiBiSage+QueryPin+PB show improvement in the recommendation performance across multiple metrics. However, the best performance on all the metrics is achieved by MultiBiSage trained with all six bipartite graphs.  The difference in the performance of MultiBiSage with all the models is statistically significant with pair t-test at significance level 0.01. In  Table \ref{tab:ablation}, we observe that training MultiBiSage on each bipartite graph alone results in poorer performance as compared to the PinSage model. This result is less surprising as the Pin-Board graph is orders of magnitude large than the other bipartite graphs.  If we train MultiBiSage with all bipartite graphs except Pin-Board, we see improvement over PinSage on Ads and Idea Pins. 

\begin{table*}[t]
\centering
\resizebox{0.9\linewidth}{!}{
\begin{tabular}{l|l|l|C|C|C|C|C}
\toprule
Query-Pin         & Model & Engaged-Pin & Rank-1 & Rank-2 & Rank-3 & Rank-4  & Rank-5 \\ 
\midrule
\multirow{2}{*}{
\includegraphics[width=0.85in]{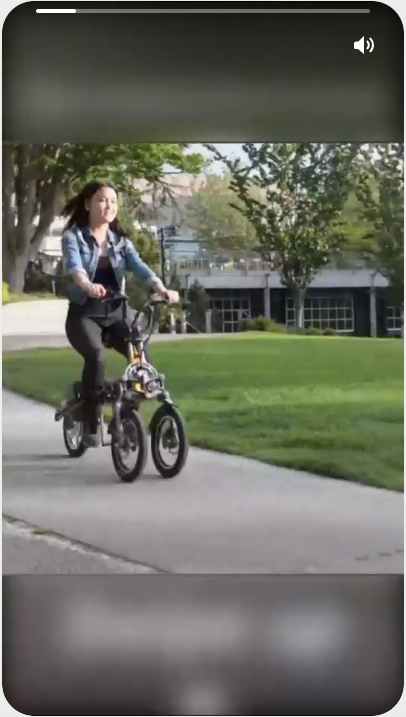}}

          &  PinSage     & 
\includegraphics[width=0.7in]{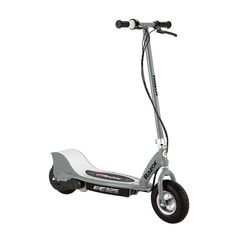}  &     \includegraphics[width=0.7in]{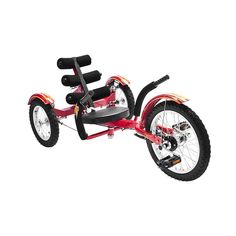}         &  \includegraphics[width=0.7in]{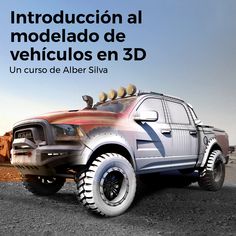}         &     \includegraphics[width=0.7in]{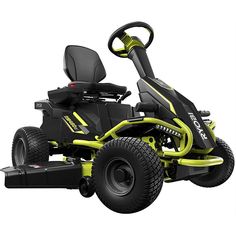}         &     \includegraphics[width=0.7in]{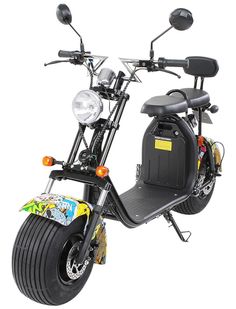}      & 
\includegraphics[width=0.7in]{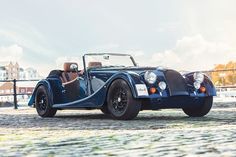}   \\
& & Rank: 364 & & & &  \\ 
\cline{2-8} \\    &   MultiBiSage    &   \includegraphics[width=0.7in]{figures/case_study/engaged.jpg}  &     \includegraphics[width=0.7in]{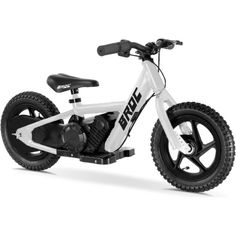}   & \includegraphics[width=0.7in]{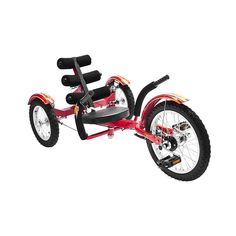}       &     \includegraphics[width=0.7in]{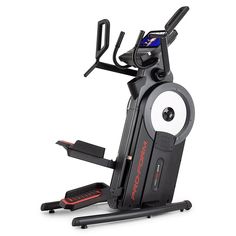}   &     \includegraphics[width=0.7in]{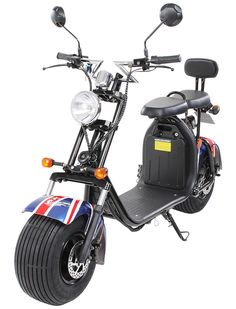}     &
         \includegraphics[width=0.7in]{figures/case_study/engaged.jpg}    \\ 
         & & Rank: 5 & & & &  \\ 
\bottomrule
\end{tabular}
}
\caption{Case study}
\label{tab:case_study}
\vspace{-1em}
\end{table*}

\subsubsection{Case Study:} We present a case study in Table \ref{tab:case_study} where a user first interacts with the shown query pin (a foldable electric scooter) and then immediately interacts with the engaged pin. The query pin is contained in a board titled "Electric Vehicles" and that board also contains cars. As a result, the top-5 recommendations provided by the PinSage model contain vehicles including cars. However, the engaged pin's embedding learned by PinSage is not too similar to that of query pin's embedding, as the rank of engaged-pin -- computed based on the similarity between the query and engaged pin's embeddings --  is 364. On the other hand, MultiBiSage learns the engaged pin's embedding closer to that of query-pin. As a result, engaged pin's rank is 5 with MultiBiSage -- a positive recall with ten nearest neighbors. In addition, we see that the other 3 nearest pins retrieved by MultiBiSage are also related to electric scooters.

%% file: conclusion.tex
We develop a novel web-scale deep learning model, called MultiBiSage, that leverages multiple bipartite graphs (each capturing diverse entities and different types of interactions) to learn high-quality pin embeddings on a production system at Pinterest. We train MultiBiSage on six bipartite graphs that include the billion nodes Pin-Board graph. Our experiments show that our proposed MultiBiSage model can significantly outperform the currently deployed PinSage production system on multiple user engagement metrics. A key novelty is how we aggregate the pin features along with the local graph structure of nodes from multiple bipartite graphs within our proposed MultiBiSage model.  We also rely on several key optimizations to ensure the model can accommodate Pinterest scale data.

%% file: appendix.tex
\appendix
\section{Pixie}
\label{algo:pixie}
The algorithm \ref{algo:randomwalk} outlines the random-walk code of Pixie. We set the number of steps per node $nw=10,000$ and random-walk reset probability $\alpha=0.9$.

\alglanguage{pseudocode}
\begin{algorithm}[t]
    \small
    \caption{Random-walk pseudocode of Pixie}
    \label{algo:randomwalk}
    \begin{flushleft}
        \textbf{Input}: Bipartite Graph $G$, number of steps per node $nw$, random-walk reset probability $\alpha$ \\
        \textbf{Output}: top-k highly visited nodes
    \end{flushleft}
\begin{algorithmic}[1]
\State Initialize Visits=\{\}
\For {each pin $p$ in $G$} 
\State current\_node=p
\For {each step $i$ in $nw$} 
\While {true}
\State Sample a neighbor node from the current\_node's neighbors by performing uniform random sampling.
\State Visits[p][current\_node]+=1 if current\_node of type pin.
\State Flip a random coin with heads probability=$\alpha$. Break if heads.

\EndWhile
\EndFor
\EndFor
\State Return top-k neighbors of each pin from Visits.
\end{algorithmic}
\end{algorithm}

\section{Experimental Setup Details}
The node features details of Pinterest dataset is summarized below
\begin{itemize}
    \item Visual: A 1024-dimensional unified visual embedding computed through a large-scale Transformer-based pretraining\cite{beal2022billion}.
    \item Textual: A 64-dimensional embedding learned by Pintext \cite{zhuang2019pintext} system.
\end{itemize}

\textbf{Training details:} For the model parameters, we use three-layer FeedForward Networks for visual features: 1024 $\rightarrow$ 2048 $\rightarrow$ 512. The visual features FFN have dropout set to 0.25 and has ReLU activations. The textual features FFN consist of 1 linear layer: 64  $\rightarrow$  512. We set the batch size to 8032 and the number of epochs to 100k. The output of the MultiBiSage model is L2-normalized 256-dimensional pin embeddings. The graph construction and feature extraction jobs execute on Spark cluster with 900 executors each with 40g executor and 40g driver memory. Pixie random-walks are performed on a single AWS x1.32xlarge (2 TB RAM, 128 cpus) instance. Model training is done in a distributed manner with two p3dn.24xlarge instances. Each p3dn.24xlarge instance has 96 custom vCPUs, 8 NVIDIA V100 Tensor Core GPUs with 32 GB of memory each.